\documentclass[letterpaper, 10 pt, journal, twoside]{IEEEtran}

\IEEEoverridecommandlockouts                              

\usepackage{epsfig} 
\usepackage{times} 
\usepackage{amsmath} 
\usepackage{amssymb}  
\usepackage{graphicx}
\usepackage{hyperref}

\usepackage{bm}
\usepackage{footnote}
\usepackage{hhline}
\usepackage{multirow}
\usepackage{subcaption}

\newcommand{\fig}[1]{Fig.~\ref{#1}}
\newcommand{\sect}[1]{Sec.~\ref{#1}}
\newcommand{\tab}[1]{Tab.~\ref{#1}}

\newcommand{\ppos}{\bm{p}}
\newcommand{\phist}{\mathcal{H}}
\newcommand{\ppred}{\mathcal{T}}
\DeclareMathOperator*{\argmax}{arg\,max}

\title{What the Constant Velocity Model Can Teach Us About Pedestrian Motion Prediction}

\author{Christoph Sch\"oller$^{1,2}$, Vincent Aravantinos$^{1}$, Florian Lay$^{1,2}$ and Alois Knoll$^{2}$
\thanks{\textcopyright2019 IEEE. Personal use of this material is permitted.  Permission from IEEE must be obtained for all other uses, in any current or future media, including reprinting/republishing this material for advertising or promotional purposes, creating new collective works, for resale or redistribution to servers or lists, or reuse of any copyrighted component of this work in other works.}
\thanks{This work was supported by the German Federal Ministry of Transport and Digital Infrastructure as part of the project Providentia.}
\thanks{$^{1}$fortiss, Research Institute of the Free State of Bavaria, Munich, Germany}%
\thanks{$^{2}$Technical University of Munich, Munich, Germany}%
\thanks{Code at \url{https://github.com/cschoeller/constant_velocity_pedestrian_motion}}%
}

\begin{document}
\maketitle



\begin{abstract}
Pedestrian motion prediction is a fundamental task for autonomous robots and vehicles to operate safely. In recent years many complex approaches based on neural networks have been proposed to address this problem. In this work we show that -- surprisingly -- a simple Constant Velocity Model can outperform even state-of-the-art neural models. This indicates that either neural networks are not able to make use of the additional information they are provided with, or that this information is not as relevant as commonly believed. Therefore, we analyze how neural networks process their input and how it impacts their predictions. Our analysis reveals pitfalls in training neural networks for pedestrian motion prediction and clarifies false assumptions about the problem itself. In particular, neural networks implicitly learn environmental priors that negatively impact their generalization capability, the motion history of pedestrians is irrelevant and interactions are too complex to predict. Our work shows how neural networks for pedestrian motion prediction can be thoroughly evaluated and our results indicate which research directions for neural motion prediction are promising in future.
\end{abstract}


\section{Introduction}
\label{sec:intro}
\IEEEPARstart{T}{he} accurate prediction of pedestrians' future motion is an essential capability for autonomous robots and vehicles to operate safely and to not endanger humans. Recently, many models based on neural networks have been proposed to address this problem \cite{zhang2019sr, sadeghian2018sophie, gupta2018social, alahi2016social, vemula2018socialatt}. Neural networks are powerful function approximators and believed to be able to take into account the pedestrians' motion histories and to learn how they interact.

\begin{figure}[t]
    \begin{center}
        \includegraphics[width=0.42\textwidth]{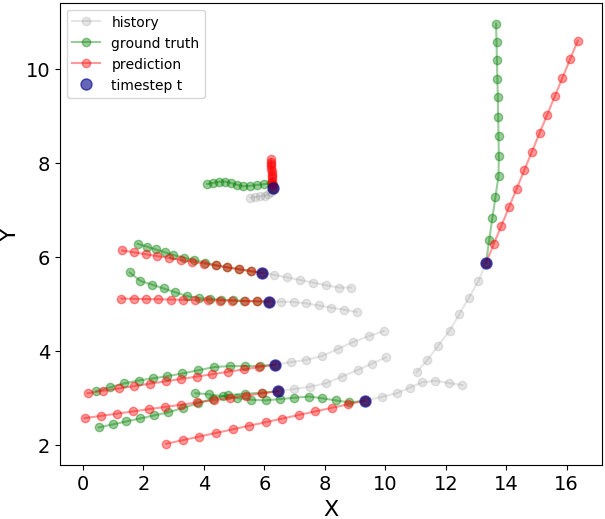}
    \end{center}
   \caption{Predictions of the Constant Velocity Model on a scene from the UCY Uni data.}
    \label{fig:predictions}
\end{figure}

In this work we show that -- surprisingly -- a simple Constant Velocity Model (CVM) can achieve state-of-the-art performance on this problem. We demonstrate this with an extensive evaluation on two well-known benchmark datasets and compare the CVM with multiple baselines and four state-of-the-art models that are based on neural networks.

Because the CVM does not require any information besides the pedestrian's last relative motion, its success indicates that either the additional information provided to neural networks is not used, or it is less relevant than commonly believed. For this reason we analyze how neural networks process their inputs and how it impacts their predictions. Our analysis reveals pitfalls in training neural networks and clarifies false assumptions about pedestrian motion prediction itself. In particular, we analyze:

\begin{itemize}
    \item \textbf{Environmental Priors}. Physical constraints and environmental semantics bias pedestrian motion towards certain patterns. We show that neural networks implicitly learn such a prior, even though no environment information has been explicitly provided to them. This learned environmental prior has a strong negative impact on their generalization to new scenes.
    \item \textbf{Motion History}. It is common belief that taking longer motion histories into account leads to more accurate predictions of the future. However, our analysis demonstrates that most of this information is redundant and consequently ignored by neural networks. Depriving a neural network of the longer past does not lead to prediction degradation.
    \item \textbf{Pedestrian Interactions}. Interactions between pedestrians happen. However, our experiments show that they are too complex to predict for neural networks and in most cases do not have a significant influence on pedestrians' trajectories. Furthermore, providing neural networks only with neighboring pedestrians' histories even negatively influences their performance.
\end{itemize}

\section{Related Work}
\label{sec:related}
The prediction of pedestrian motion has been addressed from various perspectives. In tracking algorithms, motion prediction is important for robust statistical filtering. To track people in images, Pellegrini et al. propose Linear Trajectory Avoidance models for short-term predictions~\cite{pellegrini2009eth, pellegrini2011predicting}. Baxter et al.~\cite{baxter2015headpose} extend a Kalman Filter with an instantaneous prior belief about where people will move, based on where they are currently looking at. Kooij et al.~\cite{kooij2019context} describe the motion of vulnerable road users with a Dynamic Bayesian Network. Ess et al.~\cite{ess2010object} and Leigh et al.~\cite{leigh2015lasertracking} use a Constant Velocity Model in combination with a Kalman Filter in their tracking approaches. Moreover, models that take into account grouping behavior have been explored for prediction~\cite{leal2011everybody, yamaguchi2011you}.

While tracking requires good short-term predictions, in this work we focus on long-term predictions. For long-term predictions, Becker et al.~\cite{becker2018red} use a recurrent encoder with a multilayer perceptron and achieve good results. Other contributions also take pedestrian interactions into account. To model these interactions, Helbing et al.~\cite{helbing1995social} propose the use of attractive and repulsive social forces. This approach was then extended and transferred to the prediction of pedestrians for autonomous robots~\cite{luber2010people}. Later, interaction-awareness has been integrated in neural networks. Alahi et al.~\cite{alahi2016social} train Long Short-term Memories (LSTMs)~\cite{hochreiter1997lstm} for pedestrian motion prediction and share information about the pedestrians through a social pooling mechanism. This mechanism gathers the hidden states of nearby pedestrians with a pooling grid. Xu et al.~\cite{xu2018crowd} propose the Crowd Interaction Deep Neural Network that computes the spatial affinity between pedestrians' last locations to weight the motion features of all pedestrians for location displacement prediction. Vemula et al.~\cite{vemula2018socialatt} address interaction-aware motion prediction with their Social Attention model by capturing the relative importance between pedestrians with spatio-temporal graphs. Zhang et al.~\cite{zhang2019sr} propose a state refinement module for LSTM networks that iteratively refines interaction-aware predictions for all pedestrians through a message passing mechanism. For interaction-aware predictions between heterogeneous agents, Ma et al.~\cite{ma2019trafficpredict} propose a graph-based LSTM that uses instance layers to take into account individual agents' movements and interactions, and category layers to exploit similarities between agents of the same type. To model distributions of future trajectories, generative neural networks have been used for interaction-aware pedestrian motion prediction. Sadeghian et al.~\cite{sadeghian2018sophie} use a Generative Adversarial Network (GAN)~\cite{goodfellow2014generative} that leverages the pedestrian's path history, and scene images as context with a physical and social attention mechanism. Gupta et al.~\cite{gupta2018social} propose the Social GAN with an extended social pooling mechanism that is not restricted to a limited grid around the pedestrian to predict as in~\cite{alahi2016social}. Amirian et al.~\cite{amirian2019social} use an InfoGAN for pedestrian motion prediction to avoid mode collapse.

Besides interaction-awareness, also the environment has been exploited for pedestrian motion prediction. Ballan et al.~\cite{ballan2016knowledge} extract navigation maps from birds eye images and transfer them to new scenes with a retrieval and patch matching procedure to make predictions. Jaipuria et al.~\cite{jaipuria2018transferable} propose a transferable framework for predicting the motion of pedestrians on street intersections based on Augmented Semi-Nonnegative Sparse Coding. Lee et al.~\cite{lee2017desire} predict the motion of vehicles and pedestrians by sampling future trajectory hypotheses from a Conditional Variational Autoencoder~\cite{kingma2013auto}. They select the most reasonable trajectories by scoring them based on future interactions and consider the environment by encoding an occupancy grid map with a Convolutional Neural Network. Bartoli et al.~\cite{bartoli2018context} consider environmental context by providing an LSTM with distances of the target pedestrian to static objects in space, as well as a human-to-human context in form of a grid map, or alternatively the neighbors' hidden encodings. Pfeiffer et al.~\cite{pfeiffer2018data} propose an LSTM that receives static obstacles as an occupancy grid and surrounding pedestrians as an angular grid. Aside from neural networks, also set-based methods~\cite{koschi2018set}, Gaussian Processes~\cite{habibi2018gps} and Reinforcement Learning algorithms~\cite{ziebart2009planning} have been used for predictions that take into account the pedestrians' environment.

In this work we focus on the long-term motion prediction for pedestrians with a CVM and analyze the implications of its success for learning models, specifically neural networks, and for pedestrian motion prediction itself.

\section{Problem Formulation}
\label{sec:problem}
We denote the position $(x_i^t, y_i^t)$ of pedestrian $i$ at time-step $t$ as $\ppos_i^t$. The goal of pedestrian motion prediction is to predict the future trajectory $\ppred_i = (\ppos_i^{t+1},...,\ppos_i^{t+n})$ for pedestrian $i$, taking into account his or her own motion history $\phist_i=(\ppos_i^0,...,\ppos_i^t)$. Interaction-aware motion prediction algorithms additionally use information about the motion histories $\{\phist_j : j \neq i\}$ of neighboring pedestrians that are present in the scene. The problem of finding a parametric model that estimates the future trajectory $\ppred_i$ can be formulated as
\begin{equation}
    \argmax_\theta P_\theta(\ppred_i \mid \phist_0,...,\phist_n),
\end{equation}
where $\theta$ are the model's parameters and $n$ the number of pedestrians in the scene. This problem is often converted in a sequence-to-sequence prediction problem, where the model can only observe information from the past.

In practice we do not directly predict the next positions for pedestrian $i$, but relative displacements, defined as
\begin{equation}
    \ppred_{\Delta i} = (\ppos_i^{t+1}-\ppos_i^{t},...,\ppos_i^{t+n}-\ppos_i^{t+n-1}).
\end{equation}
Predicting such residuals reduces the error margins compared with directly predicting absolute future positions. Knowing $\ppos_i^{t}$ allows us to convert $\ppred_{\Delta i}$ back to $\ppred_i$.

\section{Constant Velocity Model}
\label{sec:cvm}
Based on the assumption that the most recent relative motion $\Delta_i = \ppos_i^{t} - \ppos_i^{t-1}$ of a pedestrian is the most relevant predictor for his or her future trajectory, the CVM is a simple but effective prediction method. This means it assumes that the pedestrian will continue to walk with the same velocity and direction as observed from the latest two timesteps. Because it does not make use of any filtering, it is sensitive to measurement noise. In particular, it predicts the future trajectory for pedestrian $i$ as
\begin{equation}
    \ppred_{\Delta i} = (\Delta_i,...,\Delta_i),
\end{equation}
where the number of $\Delta_i$ is equal to the number of prediction steps.

\section{Experiments}
\label{sec:experiments}
\renewcommand{\arraystretch}{1.15}
\begin{table*}[t]
\begin{center}
\begin{tabular}{c|c|c|c|c|c|c|c|c||c|c|c}
\textbf{Metric} & \textbf{Dataset} & \textbf{ConstAcc} & \textbf{Lin} & \textbf{FF} & \textbf{LSTM} & \textbf{RED} & \textbf{SR-LSTM} & \textbf{OUR} & \textbf{SoPhie} & \textbf{S-GAN} &  \textbf{OUR-S} \\
\hhline{============}
             & ETH-Uni  & 1.35 & 0.58 & 0.67 & 0.54 & 0.60 & 0.63 & 0.58 & 0.70 & 0.59 & \textbf{0.43} \\ \cline{2-12}
             & Hotel    & 0.95 & 0.39 & 1.59 & 2.91 & 0.47 & 0.37 & 0.27 & 0.76 & 0.38 & \textbf{0.19} \\ \cline{2-12}
\textbf{ADE} & Zara1    & 0.59 & 0.44 & 0.39 & 0.34 & 0.34 & 0.41 & 0.34 & 0.30 & \textbf{0.18} & 0.24 \\ \cline{2-12}
             & Zara2    & 0.50 & 0.41 & 0.38 & 0.43 & 0.31 & 0.32 & 0.31 & 0.38 & \textbf{0.18} & 0.21 \\ \cline{2-12}
             & UCY-Uni  & 0.79 & 0.60 & 0.69 & 0.72 & 0.47 & 0.51 & 0.46 & 0.54 & \textbf{0.26} & 0.34 \\ \cline{1-12}
             \multicolumn{2}{c|}{\textbf{AVG}} & 0.84 & 0.48 & 0.74 & 0.99 & 0.44 & 0.45 & 0.39 & 0.54 & 0.32 & \textbf{0.28} \\ \cline{1-11}
\hhline{============}
             & ETH-Uni  & 3.29 & 1.11 & 1.32 & 1.04 & 1.14 & 1.25 & 1.15 & 1.43 & 1.04 & \textbf{0.80} \\ \cline{2-12}
             & Hotel    & 2.41 & 0.81 & 3.12 & 6.07 & 0.94 & 0.74 & 0.51 & 1.67 & 0.79 & \textbf{0.35} \\ \cline{2-12}
\textbf{FDE} & Zara1    & 1.50 & 0.93 & 0.81 & 0.74 & 0.76 & 0.90 & 0.76 & 0.63 & \textbf{0.32} & 0.48 \\ \cline{2-12}
             & Zara2    & 1.30 & 0.83 & 0.77 & 0.95 & 0.70 & 0.70 & 0.69 & 0.78 & \textbf{0.34} & 0.45 \\ \cline{2-12}
             & UCY-Uni  & 2.03 & 1.19 & 1.38 & 1.60 & 1.00 & 1.10 & 1.02 & 1.24 & \textbf{0.49} & 0.71 \\ \cline{1-12}
             \multicolumn{2}{c|}{\textbf{AVG}} & 2.11 & 0.97 & 1.48 & 2.08 & 0.91 & 0.94 & 0.83 & 1.15 & 0.60 & \textbf{0.56} \\
\hhline{============}
\end{tabular}
\end{center}
\caption{Displacement errors of the CVM, common baselines and state-of-the-art models.}
\label{tab:experiments}
\end{table*}

\begin{figure*}[t]
    \centering
    \begin{subfigure}[b]{0.33\textwidth}
        \includegraphics[width=0.98\textwidth]{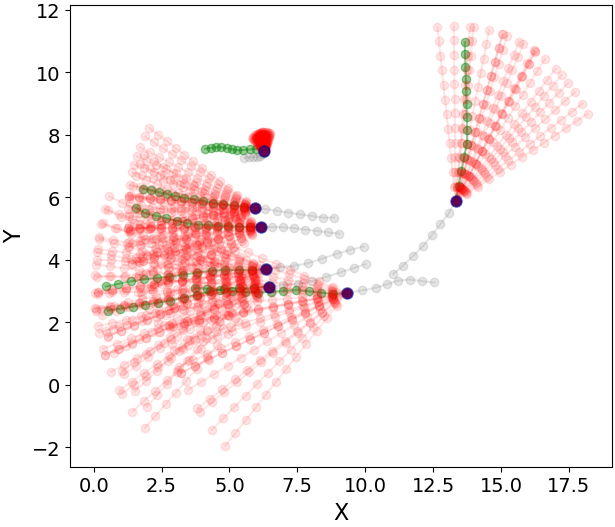}
        \caption{OUR-S}
        \label{fig:pred-cvm-s}
    \end{subfigure}%
    \begin{subfigure}[b]{0.33\textwidth}
        \includegraphics[width=0.98\textwidth]{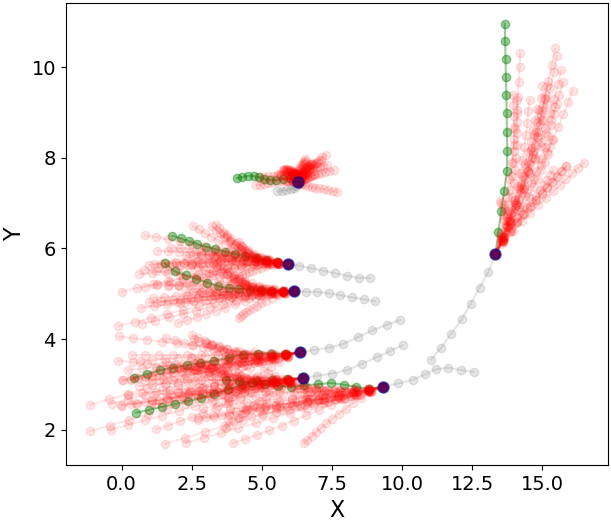}
        \caption{S-GAN}
        \label{fig:pred-sgan}
    \end{subfigure}%
        \begin{subfigure}[b]{0.33\textwidth}
        \includegraphics[width=0.98\textwidth]{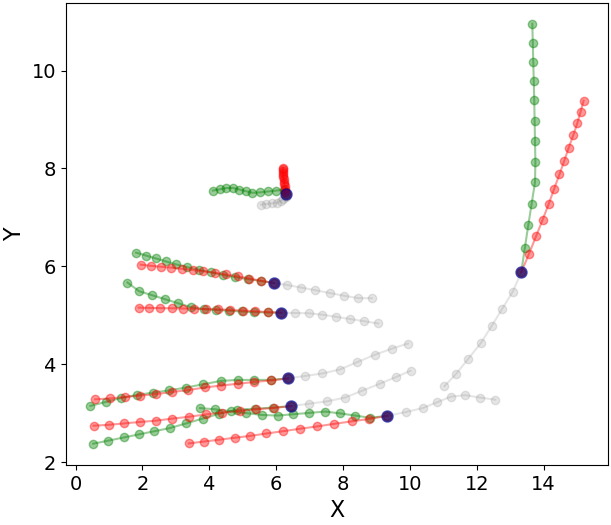}
        \caption{RED}
        \label{fig:pred-red}
    \end{subfigure}%
    \caption{The same scene predicted with OUR-S, S-GAN and RED. For OUR-S and S-GAN we drew 20 samples as explained in~\sect{sec:models}. Even though the pedestrians' trajectories are non-linear, the CVM generates close approximations.}
    \label{fig:qualitative-comp}
\end{figure*}

To evaluate the performance of the CVM and compare it with other models, we use two datasets for pedestrian motion prediction, the ETH~dataset~\cite{pellegrini2009eth} and the UCY~dataset~\cite{lerner2007ucy}. In total they contain 1950 unique pedestrians and five scenes:~ETH-Uni and Hotel (from ETH) and Zara1, Zara2 and UCY-Uni (from UCY). Each scene has a different walkway layout and contains non-traversable obstacles. The datasets are based on real-world video recordings from different scenarios, like university campuses and pedestrian walking zones. Both datasets have been heavily used to evaluate the performance of motion prediction models in recent contributions \cite{zhang2019sr, sadeghian2018sophie, gupta2018social, alahi2016social, vemula2018socialatt, becker2018red, xu2018crowd, amirian2019social}.

In our experiment we observe the last 8 positions of a pedestrian's trajectory and predict the next 12 timesteps. This corresponds to an observation window of 3.2 seconds and a prediction for the next 4.8 seconds, which is an established setting and used in other motion prediction papers as well \cite{zhang2019sr, sadeghian2018sophie, gupta2018social, alahi2016social, vemula2018socialatt, becker2018red, amirian2019social}. This means we slice trajectories with a sliding window and step-size of one into sequences of length 20. We reject sliced trajectories with a length shorter than 10. By this we guarantee an observation of 8 timesteps and that the evaluated models must predict at minimum the next two timesteps.

As proposed in~\cite{alahi2016social}, we train on four scenes and evaluate on the remaining one in a leave-one-out cross-validation fashion. This ensures the evaluation of the model's generalization capability to new scenarios.

Like in related contributions we report errors in meters and evaluate all models with the following metrics:
\begin{itemize}
    \item \textit{Average Displacement Error} (\textbf{ADE}) --- Average L2 distance between all corresponding positions in ground truth and predicted trajectory.
    \item \textit{Final Displacement Error} (\textbf{FDE}) --- L2 distance between the last position in ground truth and the last position in predicted trajectory.
\end{itemize}

\subsection{Training}
\label{sec:training}
We trained each model -- except state-of-the-art models and the CVM that does not require training -- with the Adam Optimizer~\cite{kingma2014adam} with learning rate 0.0004, batch size 64, and for 35 epochs. All hyperparameter were determined empirically. As loss function we used the \textit{Mean Squared Error}. We randomly split the training scenes into a training set and a 10\% validation set to detect overfitting. All models converged without overfitting and did not require further hyperparameter tuning. State-of-the-art models from other contributions were trained as described in respective papers, including specified data augmentations and loss functions.

\subsection{Models}
\label{sec:models}
We compare the performance of the CVM with multiple baselines that are commonly used in contributions to the pedestrian motion prediction domain:

\begin{itemize}
    \item \textit{Constant Acceleration} (\textbf{ConstAcc}) --- Observes the last three positions of a pedestrian and assumes he or she continues to walk with the same acceleration.
    \item \textit{Linear Regression} (\textbf{Lin}) --- Multivariate multi-target linear regression model that estimates each component in the predicted trajectory as an independent linear regression. Each predicted variable depends on the full motion history.
    \item \textit{Feed Forward Neural Network} (\textbf{FF}) --- Fully connected neural network that receives all eight motion history timesteps as a flattened vector. It then applies two hidden layers with 60 and 30 neurons, respectively. Both hidden layers are followed by ReLu activations. The final linear output layer has 24 outputs, which corresponds to 12 prediction timesteps. We also evaluated bigger networks for this comparison and following analysis section. This lead to worse performance in this comparison (stronger overfitting) and the same conclusions in the analysis.
    \item \textit{LSTM Network} (\textbf{LSTM}) --- Stacked LSTM that receives single positions and linearly embeds them in a 32 dimensional vector. Three LSTM layers with 128 hidden dimensions and a linear output layer follow. We trained the LSTM with Teacher Forcing~\cite{williams1989teacher}.
\end{itemize}

We further include four state-of-the-art models in our evaluation. These are the \textit{RNN-Encoder-MLP} (\textbf{RED})~\cite{becker2018red}, the \textit{LSTM with State Refinement} \textbf{SR-LSTM}~\cite{zhang2019sr} and for generative models \textit{Social GAN} (\textbf{S-GAN})~\cite{gupta2018social} and the \textit{SoPhie GAN} (\textbf{SoPhie})~\cite{sadeghian2018sophie}. We denote the CVM as \textbf{OUR}.

Because S-GAN and SoPhie were evaluated by drawing 20 samples and taking the predicted trajectories with minimum ADE and FDE for evaluation into account, we add an extended version \textbf{OUR-S} of the CVM for comparability. For OUR-S we add additional angular noise to its predicted direction, which we draw from $\mathcal{N}(0,\sigma^2)$ with $\sigma = 25^{\circ}$ and evaluate its error in the same fashion.

While for SR-LSTM and SoPhie GAN we report the original papers' results, for S-GAN we use a pre-trained\footnote{github.com/agrimgupta92/sgan} version of their best performing model, which is not interaction-aware as reported in~\cite{gupta2018social}. Note this is an improved version compared to what the authors report in their paper. To evaluate RED, we carefully re-implemented and trained the model as proposed by the authors. 

\subsection{Results}
\label{sec:results}
\textbf{Quantitative}. In~\tab{tab:experiments} we display the prediction errors for all evaluated models on each scene. On average the best performing model for both ADE and FDE was OUR-S, which is the CVM with sampling. It also outperformed state-of-the-art generative models S-GAN and SoPhie. As explained in the previous section, OUR-S, S-GAN and SoPhie were evaluated by considering only the errors of the best predicted samples. For this reason, the other models are discussed separately.

Of the models without sampling, OUR outperformed all other models as well, including state-of-the-art RED and SR-LSTM. It's advantage was especially strong for the Hotel scene. Among the basic neural networks FF and LSTM, FF outperformed LSTM. We hypothesize that an error accumulation effect for the LSTM could be responsible for this, as it predicts the next step based on its output for the last step. ConstAcc performed the worst, which shows that especially over long prediction horizons the assumption of continual acceleration or deceleration is detrimental. To our surprise, Lin performed better than LSTM, compared to what~\cite{vemula2018socialatt} reported. We believe this discrepancy can be attributed to the data augmentation they used for all their models. The good performance of Lin can be explained with the high bias and thus strong generalization of linear regression models.

\textbf{Qualitative}. \fig{fig:predictions} shows predictions of OUR on UCY Uni. The worst prediction OUR made is for the pedestrian at the top left, who suddenly makes a sharp turn. This behavior is not predictable based on a pedestrian's motion history. The rightmost pedestrian walks in a gradual curve. As the CVM makes linear predictions it can not extrapolate this behavior. However, often these trends are not reliable and the trajectory curvature suddenly changes. For example, the bottommost pedestrian is first taking a turn to the left, but then changes his or her trajectory back to the right, which is difficult to foresee. Overall, the linear predictions of the CVM are good approximations of the pedestrians' future trajectories.

In~\fig{fig:qualitative-comp} we show predictions of OUR-S, S-GAN and RED for the same scene as in~\fig{fig:predictions}. For OUR-S and S-GAN we sampled 20 trajectories as described in~\sect{sec:models}. The width of the prediction cone of OUR-S in~\fig{fig:pred-cvm-s} can be controlled with $\sigma$ during sampling. OUR-S is able to predict accurate linear approximations of the pedestrians' future trajectories, even for those pedestrians that walk in a curved fashion. The predictions of S-GAN in~\fig{fig:pred-sgan} appear to be clustered into maneuvers, like walking straight, turning right, or left. However, the pedestrians don't always seem to follow these patterns. The predictions of RED in~\fig{fig:pred-red} are approximately linear in this scene, which confirms that this is a good strategy.

\section{Analysis of Neural Network Behavior}
\label{sec:analysis}
Our experiments show that the CVM performs strong on pedestrian motion prediction, despite its simplicity and lack of provided information. This indicates that the neural networks it outperforms are either not able to use the additional information they are provided with to their advantage, or that this information is not as relevant as commonly believed. In this section we develop explanations for this result. In particular, we analyze the influence of environmental priors, the motion history and pedestrian interactions on neural network performance. For our analysis we use the feed forward neural network (FF) and the RNN-Encoder-MLP (RED)~\cite{becker2018red} from~\sect{sec:models}. Both models have simple training dynamics. This enables us to disentangle the effects of our experiments on the models' performance from noise, caused by instabilities and randomness in the training process. While FF is a basic neural network, RED is a state-of-the-art model with good performance.

\begin{figure}[t]
    \begin{center}
        \includegraphics[width=0.42\textwidth]{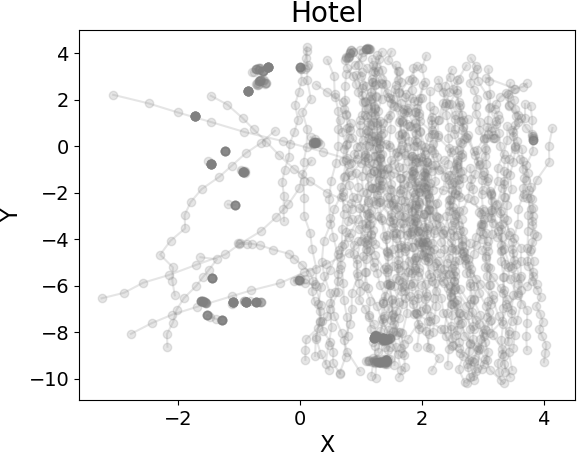}\\
        \vspace{0.3cm}
        \includegraphics[width=0.42\textwidth]{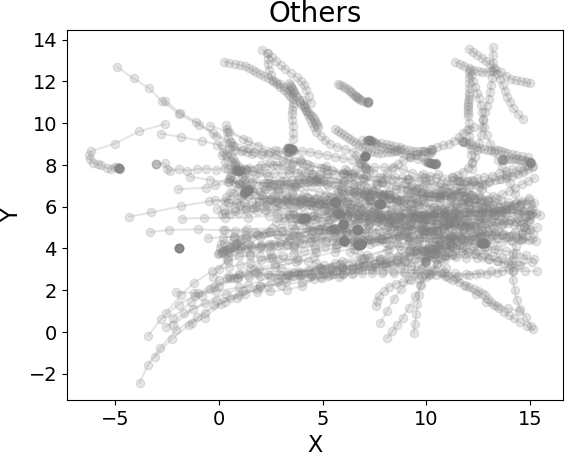}
    \end{center}
    \caption{Trajectories from scene Hotel and other scenes combined. We sub-sampled the datasets for better visibility. The majority of trajectories in Hotel are oriented vertically, whereas in the other scenes most trajectories run horizontally. This causes the learning of an environmental prior which contradicts the Hotel scene at test time.}
    \label{fig:environment}
\end{figure}

\subsection{Environmental Priors}
\label{sec:environment}
The environment of each scene puts constraints and bias on how pedestrians move within it. Constraints can be physical, for example certain areas like buildings cannot be traversed. Bias can be caused by the semantics of a scene, e.g. on a parking lot of a shopping center people are likely to either walk towards the shopping center, or away from it.

We argue that even though this environmental information was not explicitly provided to the networks in \sect{sec:experiments}, they implicitly learn such prior. Each area of a scene corresponds to a specific numeric range of input coordinates. This allows a neural network to associate these areas with certain motion patterns. But even if we prevent the network from making this association, it would still learn motion patterns that are typical for the whole scene.

\begin{table}[b]
\begin{center}
\begin{tabular}{c|c|c|c|c}
\textbf{Model} & \textbf{Metric} & \textbf{Basic} & \textbf{Relative} & \textbf{Rotations} \\
\hhline{=====}
\multirow{4}{*}{FF} & \textbf{ADE Hotel} & 1.59 & 0.45 & \textbf{0.30} \\ \cline{2-5}
 & \textbf{FDE Hotel} & 3.12 & 0.95 & \textbf{0.55} \\ \cline{2-5}
 & \textbf{ADE Avg} & 0.74 & 0.44 & \textbf{0.42} \\ \cline{2-5}
 & \textbf{FDE Avg}   & 1.48 & 0.93 & \textbf{0.87} \\ 
 \hhline{=====}
 \multirow{4}{*}{RED} & \textbf{ADE Hotel} & 1.59 & 0.45 & \textbf{0.30} \\ \cline{2-5}
 & \textbf{FDE Hotel} & 2.80 & 0.92 & \textbf{0.56} \\ \cline{2-5}
 & \textbf{ADE Avg} & 0.79 & 0.44 & \textbf{0.41} \\ \cline{2-5}
 & \textbf{FDE Avg}   & 1.49 & 0.92 & \textbf{0.86} \\
\hhline{=====}
\end{tabular}
\end{center}
\caption{Effects of learning environmental priors.}
\label{tab:environment}
\end{table}

To verify this hypothesis, we compare the models FF and RED from \sect{sec:models} that we train without data augmentations with two modifications that dampen the effects of learning environmental priors. For the first modification (\textbf{Relative}) we do not feed the network with absolute positions as inputs, but instead with the pedestrian's past relative motion, which we compute analog to $\ppred_{\Delta i}$ in~\sect{sec:problem}. This ensures that the model can not learn to associate certain areas in the scene with a specific motion pattern. For the second modification (\textbf{Rotations)}, we use relative motion and additionally add random rotations to the training trajectories to reduce directional bias. For this purpose we sample angles from $\mathcal{N}(0,\sigma^2)$ with $\sigma = 180^{\circ}$. Note that we only apply one rotation to each sample in the training dataset, such that the resulting dataset has the same size for all three training variations. This ensures a fair comparison.

The ADE and FDE for each variant are displayed in \tab{tab:environment}. For the rows with \textbf{Avg} we report the average errors of the leave-one-out experiment as in \sect{sec:experiments}. Our results show that Relative, as well as the additional Rotations strongly improved the models' performance. This effect is even intensified for scene \textbf{Hotel}, which we therefore report additionally. To understand why Hotel was stronger influenced by learning environmental priors, we plotted a sub-sample of the trajectories of the Hotel scene, as well as of all other scenes combined in \fig{fig:environment}. It shows that most trajectories in the other scenes run horizontally, whereas in Hotel most trajectories run vertically. This explains our observation and we conclude that the networks learned this motion pattern as a prior that negatively influences their generalization capability.

While the benefits of data augmentation have been reported earlier \cite{sadeghian2018sophie,becker2018red}, our analysis shows that the reason for this is that it helps to prevent learning environmental priors. The awareness of this problem is also necessary to understand certain phenomena, for example, we hypothesize that learning environmental priors is the primary reason for the bad performance of LSTMs reported in~\cite{vemula2018socialatt}, instead of the missing interaction-awareness as the authors suggest. To train the LSTM the authors used absolute positions and applied no data augmentation. 

\subsection{Motion History}
\label{sec:motion-history}
It is believed that neural networks can use long motion histories to make better predictions. In~\sect{sec:experiments} we provided our models with a history of eight timesteps, which is common practice in the domain of pedestrian motion prediction \cite{zhang2019sr, sadeghian2018sophie, gupta2018social, alahi2016social, vemula2018socialatt, becker2018red, amirian2019social}. However, the performance of the CVM that only uses the last two timesteps to make predictions suggests that for pedestrian motion prediction long histories are not as relevant as believed. To isolate the effects of learning environmental priors and the motion history, we use the data augmentations from \sect{sec:environment}.

To evaluate if the models utilize the full motion history to make predictions, we compute gradient norms for each timestep with respect to the predicted trajectory. In particular, after training we keep the network static and summarize predicted trajectories in a scalar value by summing up the absolute values of the network's predicted displacements. As our network is a function $f(\phist_i)$ of the motion history, we can compute gradients
\begin{equation}
        \triangledown_i^t f = \left[ \frac{\mathrm{d} f(\phist_i)}{\mathrm{d} x_i^t}, \frac{\mathrm{d} f(\phist_i)}{\mathrm{d} y_i^t} \right]
\end{equation}
for each timestep $t$ in history $\phist_i$. Then we compute the norm $\lVert \triangledown_i^t f \rVert$ of each gradient and evaluate it for all testset trajectories with respectively trained models. We sum the gradient norms for all trajectories per timestep and normalize the resulting values to a distribution, such that they sum up to one. This distribution expresses how much influence each timestep in motion history $\phist_i$ has on average on the output trajectory. 

We found that for model FF the latest relative motion has an influence of 68.2\% on the predicted trajectory, while timestep $t-1$ contributes only 8.1\%. The other 5 timesteps in the relative motion history influenced the prediction with 3.8\% -- 6\% and their influence did not decrease monotonically, but fluctuated. This fluctuation would be counter-intuitive if all timesteps contain predictive information with decreasing relevance. For model RED the influence of the latest timestep was 80.3\% and timestep $t-1$ only 8.3\% and thus the influence drop is even stronger. This is likely because of the recurrent encoder of RED. In summary, the non-monotonic fluctuations of earlier timesteps' influence, combined with the strong drop from timestep $t$ to $t-1$ indicates that earlier timesteps in the history are not predictive.

\begin{figure}[ht]
    \centering
    \includegraphics[width=0.31\textwidth]{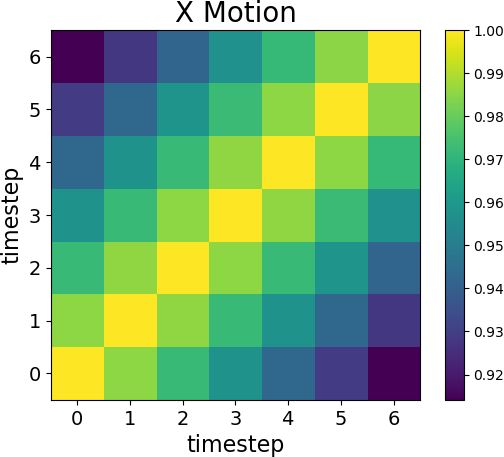}
    \par\bigskip
    \includegraphics[width=0.31\textwidth]{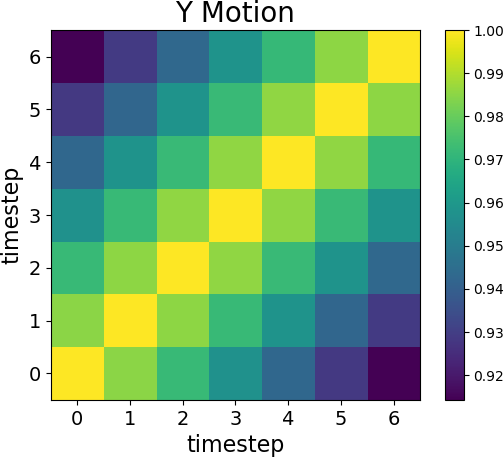}
    \caption{Pearson correlation coefficients between all timesteps of the observed relative motion histories for X and Y. All timesteps are highly correlated.}
    \label{fig:hist-correlations}
\end{figure}

We further computed linear correlation coefficients between all timesteps in the history for the $X$ and $Y$ components. The resulting correlations are displayed in~\fig{fig:hist-correlations}. All timesteps are highly correlated, with correlation coefficients ranging from 0.91 to 1.0. The closer the timesteps are, the more they are correlated. While some correlation of positions in the motion history is to be expected, this high correlation means that the timesteps contain mostly shared and thus redundant information. High correlation of features is an undesirable property in regression problems such as motion prediction. For the neural networks we evaluated, this redundant information likely acts as noise rather than signal. This explains why the networks strongly rely on the latest timestep for prediction.

To empirically confirm our findings we analyzed if the performance of the neural networks deteriorates by successively depriving them more of the pedestrian's motion history. In particular, we train the networks with relative histories of sizes $7,...,1$, but let them always predict the next 12 timesteps as usual.
\begin{table}[t]
\begin{center}
\begin{tabular}{c|c|c|c||c}
\textbf{Model} & \textbf{Metric} & \textbf{Full History} & \textbf{History Size One} & $\bm{\sigma}$ \\
\hhline{=====}
\multirow{2}{*}{FF} & \textbf{ADE Avg} & 0.42 & \textbf{0.41} & 0.002 \\ \cline{2-5}
 & \textbf{FDE Avg}   & 0.87 & \textbf{0.86} & 0.005 \\ 
 \hhline{=====}
 \multirow{2}{*}{RED} & \textbf{ADE Avg} & 0.41 & \textbf{0.40} & 0.003 \\ \cline{2-5}
 & \textbf{FDE Avg}   & 0.86 & \textbf{0.84} & 0.005 \\
\hhline{=====}
\end{tabular}
\end{center}
\caption{Depriving the networks of motion history.}
\label{tab:history}
\end{table}
\tab{tab:history} shows the networks' performance when providing them with the full motion history and a history of size one. We further evaluated all intermediate history sizes and the performance of the models stayed approximately the same for all settings. There was no monotonic decrease in performance during history deprivation, but the performance fluctuated. We report the sample standard deviations of these fluctuations in \tab{tab:history} as well. These small fluctuations can be attributed to randomness in the network's training process, i.e. weight initialization and stochastic gradient descent. Our findings confirm that contrary to popular belief, for pedestrian motion prediction a long motion history is not predictive for the future.

\subsection{Pedestrian Interactions}
\label{sec:interactions}
To behave in an interaction-aware manner, a person must anticipate the future motion of his neighbors. Only then it is possible to plan his or her trajectory such that potential collisions are avoided. This implies that a neural network that predicts an interaction-aware trajectory for a pedestrian~$i$ must implicitly and simultaneously predict the future trajectories of the pedestrian's neighbors. The CVM does not receive information about surrounding pedestrians and is not making interaction-aware predictions. Its strong performance hints that either interactions do not have a strong influence on the pedestrians' trajectories, or the state-space of possible interactions is too big and complex for neural networks to make robust predictions that reliably decrease the expected error.

To analyze this, we compare three variations of training our models FF and RED. For the experiments in this section we modify RED and provide its decoder with additional information about the target pedestrians neighbors by concatenation to the target encoding. We again apply the data augmentations from~\sect{sec:environment} for both models. In the first version (\textbf{Basic}), the models do not receive any neighborhood information. In the second version (\textbf{History}), the models receive the last eight history steps --~including~timestep~$t$~-- of the pedestrian's 12 closest neighbors. Note that also state-of-the-art models like \cite{gupta2018social}, \cite{alahi2016social} or \cite{vemula2018socialatt} receive information about the neighbors' past, but indirectly through specialized pooling modules. In the third version (\textbf{Future}), the models are provided with 12 true future positions of the pedestrian's 12 closest neighbors, starting from timestep $t+1$. They have perfect information about the neighbors' future trajectories and should be able to utilize it if this information is relevant for making predictions. We choose 12 as the number of observed neighbors, because the average number of neighbors across all scenes is 11.34 and provide all neighbor positions relative to the position $\ppos^t_i$ of the target pedestrian $i$. We order the observed neighbors by their distance to pedestrian $i$ at timestep $t$ in ascending order and pad missing neighbors with zeros. We do not include neighbors with partial trajectories in the observed time window, as this has a negative impact on the prediction performance of History and Future. We re-train each model variant separately as in \sect{sec:motion-history}.

\begin{figure}[t]
    \begin{center}
        \includegraphics[width=0.42\textwidth]{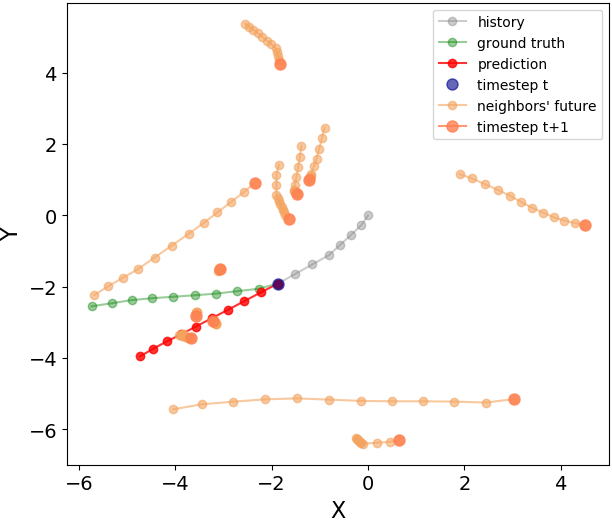}
    \end{center}
   \caption{Prediction of the model FF that received the neighbors' future trajectories as input. Despite this additional information it predicted a colliding trajectory.}
    \label{fig:interaction-future}
\end{figure}

\tab{tab:interactions} shows the results of our experiments for all three training variants. Both models performed the worst when they received information about the neighbors histories. This is likely because pedestrian interactions are too complex and the model is not able to determine robust solutions while internally predicting for all pedestrians simultaneously. Instead of being useful, in this case the additional information about the neighbors' histories rather acts as noise. Surprisingly, even the models we provide with the neighbors' future trajectories do not outperform those that receive no neighborhood information at all. They are rather on par and for RED the version Basic performs even slightly better. These results explain why the CVM can perform well without taking into account any information about the pedestrian's neighbors. The fact that none of the models is able to exploit the neighbors' true future trajectories shows that interactions between pedestrians are either not predictable, or that the average impact of interactions on the pedestrians' trajectories and thus the prediction error is small.

\begin{table}[t]
\begin{center}
\begin{tabular}{c|c|c|c|c}
\textbf{Model} & \textbf{Metric} & \textbf{Basic} & \textbf{History} & \textbf{Future} \\
\hhline{=====}
\multirow{2}{*}{FF} & \textbf{ADE Avg} & \textbf{0.42} & 0.47 & 0.44 \\ \cline{2-5}
& \textbf{FDE Avg} & 0.87 & 0.93 & \textbf{0.86} \\
\hhline{=====}
\multirow{2}{*}{RED} & \textbf{ADE Avg} & \textbf{0.41} & 0.45 & 0.44 \\ \cline{2-5}
& \textbf{FDE Avg} & \textbf{0.86} & 0.93 & 0.90 \\
\hhline{=====}
\end{tabular}
\end{center}
\caption{Influence of neighborhood information.}
\label{tab:interactions}
\end{table}

\fig{fig:interaction-future} shows a failed prediction of model FF that was trained with Future. The model predicted a path that would cause a collision with the standing neighbors despite fully observing them. Besides such failures, we also observed predictions that could be interpreted as interaction-aware, but these were so infrequent that rather chance and not interaction-awareness caused them. Furthermore, most true trajectories do not involve obvious interaction-aware behavior, which makes observing interaction-awareness additionally difficult.

Our analysis indicates that interactions between pedestrians are less relevant than commonly believed for making accurate predictions. Furthermore, interaction-aware predictions are too complex to solve only based on neighbors' motion histories. Our results are consistent with the observations made by \cite{gupta2018social}, that including interaction-awareness by providing the model with the neighbors' motion histories does not lead to performance gains and can even be detrimental.

\section{Conclusion}
\label{sec:conclusion}
In this work we have shown with extensive experiments, that the CVM can outperform state-of-the-art models in pedestrian motion prediction. Inspired by this result, we analyzed why neural networks can not use their computational power and the additional information they are provided with to their advantage. We found out that neural networks learn environmental priors that negatively influence their generalization and explained how data augmentation can help to alleviate this problem. Furthermore, we showed that the long motion history of a pedestrian is not relevant for making predictions and it mostly contains redundant information that neural networks ignore. Lastly, our analysis indicates that interactions between pedestrians are less relevant than commonly believed. Based on the neighbors' motion histories interactions are too complex to reliably predict and likely do not have a significant influence on pedestrians' trajectories in most cases.

Due to its strong performance on commonly accepted benchmarks and its simplicity, the CVM should in future be included as a standard baseline for pedestrian motion prediction. Establishing strong baselines is important for advancing research, especially given a trend towards complex models~\cite{lipton2018troubling}. Furthermore, simple methods such as the CVM help to establish a better understanding of the problem at hand. This has been demonstrated in other domains, such as image classification~\cite{brendel2018approximating} or captioning~\cite{devlin2015exploring} as well.

Our results suggest that in future a stronger focus on the pedestrians' environment is a promising research direction. The development of new datasets with more diverse environments could be one way to train models that generalize better and are able to exploit environment information. Furthermore, it would be interesting to see which of our insights are transferable to other motion prediction domains. Interactions between vehicles, or pedestrians and vehicles, may be much more predictable as vehicles move in highly structured environments.

\addtolength{\textheight}{-12cm}  


\bibliographystyle{IEEEtran}
\bibliography{include/references}

\end{document}